\documentclass{article}
\usepackage{spconf,amsmath,graphicx}

\usepackage{graphicx}
\usepackage{amssymb}
\usepackage{booktabs}

\usepackage{times}
\usepackage{latexsym}
\usepackage{multicol}
\usepackage{multirow}
\usepackage[sort, numbers]{natbib}
\usepackage{graphicx}
\usepackage{microtype}
\usepackage{amssymb}
\usepackage{algorithm2e}
\usepackage{color}
\usepackage{colortbl}
\usepackage{xcolor}
\usepackage{floatrow}
\usepackage{epsfig}
\usepackage{bbding}
\usepackage{float}
\floatstyle{plaintop}
\usepackage{subcaption}
\usepackage{hyperref}
\usepackage{orcidlink}

\title{PhysID: Physics-based Interactive Dynamics from a Single-view Image}

\name{
    \normalsize
    \begin{tabular}[t]{c}
        Sourabh Vasant Gothe \orcidlink{0000-0003-4737-2218}, Ayon Chattopadhyay \orcidlink{0009-0000-4709-150X}, Gunturi Venkata Sai Phani Kiran \orcidlink{0009-0006-8761-3855}, Pratik, Vibhav Agarwal \orcidlink{0000-0002-2029-9885},\\
        Jayesh Rajkumar Vachhani \orcidlink{0000-0003-0267-4474}, Sourav Ghosh \orcidlink{0000-0003-1866-1408}, Parameswaranath VM \orcidlink{0009-0005-9784-6196}, Barath Raj KR \orcidlink{0000-0003-0451-2452}
    \end{tabular}
}
\address{
    \begin{tabular}[t]{c}
        \textit{Samsung R\&D Institute India - Bangalore}\\
        {\small\texttt{\{ sourab.gothe, a.chattopadh, g.kiran, pratik.4, vibhav.a,}}\\
        {\small\texttt{jay.vachhani, sourav.ghosh, nath.vm, barathraj.kr} \} @samsung.com}
    \end{tabular}
}

\begin{document}

\maketitle

\begin{abstract}
Transforming static images into interactive experiences remains a challenging task in computer vision. Tackling this challenge holds the potential to elevate mobile user experiences, notably through interactive and AR/VR applications. Current approaches aim to achieve this either using pre-recorded video responses or requiring multi-view images as input.  In this paper, we present PhysID, that streamlines the creation of physics-based interactive dynamics from a single-view image by leveraging large generative models for 3D mesh generation and physical property prediction. This significantly reduces the expertise required for engineering-intensive tasks like 3D modeling and intrinsic property calibration, enabling the process to be scaled with minimal manual intervention. We integrate an on-device physics-based engine for physically plausible real-time rendering with user interactions. PhysID\footnote{Project page: https://physid.github.io/} represents a leap forward in mobile-based interactive dynamics, offering real-time, non-deterministic interactions and user-personalization with efficient on-device memory consumption. Experiments evaluate the zero-shot capabilities of various Multimodal Large Language Models (MLLMs) on diverse tasks and the performance of 3D reconstruction models. These results demonstrate the cohesive functioning of all modules within the end-to-end framework, contributing to its effectiveness.\looseness=-1
\end{abstract}

\begin{keywords}
Multimodal Large Language Models, 3D Modelling, Physics-Based Rendering, Diffusion
\end{keywords}

\section{Introduction}
\label{sec:intro}

In the current era of smartphones with abundant computing capabilities, users are increasingly drawn to engage with objects within images rather than merely perceiving static scenes. Applications such as interactive wallpapers on mobile phones represent a unique blend of aesthetics and functionality, allowing users to engage with dynamic content directly from their home screens. Based on organizational big data analytics on a sample of mobile users, a significant 76.2\% now favor personalized wallpapers, and the adoption of live wallpapers has increased by 30\%, highlighting user interest in leveraging smartphone displays for more immersive and interactive experiences. However, the current state of these applications is largely constrained to pre-defined animations and lacks the sophistication to emulate the natural dynamics observed in the real world.\looseness=-1

\begin{figure}[t]
\centering		
	{\includegraphics[width=1\linewidth]{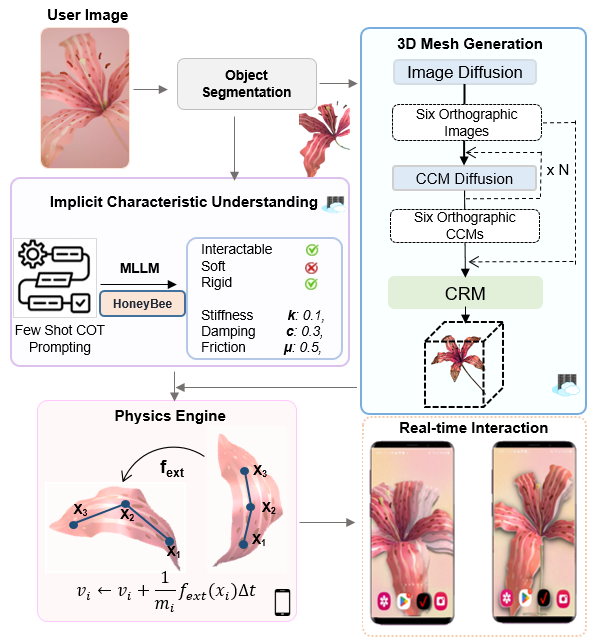}}    
	\caption{Overview of Physics-based Interactive Dynamics}
	\label{fig:architectureTeaser}
\end{figure}

Current approaches tackling interactive image dynamics diverge into distinct paths. Video-generation techniques, such as \cite{li2024generative,  shi2024motion, wu2024draganything, zhang2024physdreamer}, offer a degree of interactivity; but, they are constrained by their limited interaction scope and lack of capabilities for real-time engagement. 3D reconstruction methods \cite{xie2024physgaussian, li2023pac, petitjean2023modalnerf} have achieved realistic simulations by incorporating physics through MPM \cite{jiang2015affine} based methods. Their reliance on pre-recorded multi-view motion data for input remains a significant dependency that restricts their application.\looseness=-1 

Inspired by the capabilities of the large generative models \cite{liu2023improvedllava, cha2024honeybee, wang2023cogvlm, wang2024crm, xu2024instantmesh, liu2024one}, this paper proposes a novel perspective to simulate image dynamics. We propose PhysID, a training-free framework that leverages the reconstruction capability of  Large Reconstruction Models (LRM), the visual understanding and reasoning ability of MLLMs, and an on-device physics-based rendering engine to simulate the accurate motion synthesis on user interaction. This approach effectively bypasses the resource-intensive requirements of traditional training methods, allowing for the deployment of advanced frameworks in real-world applications without the need for extensive data annotation.\looseness=-1

As shown in Fig. \ref{fig:architectureTeaser}, PhysID processes a static image the user provides, first determining its suitability for interaction through the MLLM. Then discern the required motion dynamics and the physical attributes of the object. Subsequently, a 3D model is constructed utilizing LRM. Finally, this 3D model and the physical properties are integrated into an on-device physics-based simulation engine to facilitate interactive rendering. This work focuses on modeling the dynamic responses of two commonly encountered material categories: 1) Rigid-body dynamics, such as flowers, plants, etc. exhibiting high Young's modulus and negligible plastic strain under external loads, and 2) Soft-body dynamics, such as clothes, flags, etc. that allows for realistic deformation in response to forces to simulate the properties of materials. Our contributions are multi-fold,\looseness=-1

\begin{enumerate}
\setlength\itemsep{-0.4em}
\item An end-to-end training-free pipeline for simulating physically plausible interactions from a single-view image
\item We curate and prepare a dataset of interactable objects carefully chosen from Objaverse \cite{deitke2024objaverse}
\item We demonstrate the visual capability of MLLMs in accurately understanding the dynamics of the object.\looseness=-1
\item PhysID provides immersive, personalized, and realistic experiences to mobile users via touch-based interaction.\looseness=-1
\end{enumerate}

\section{Method}

PhysID comprises several interconnected modules that work in tandem to identify interactable elements within an image, reconstruct their 3D mesh representation, understand their physical properties, and simulate realistic user interactions.\looseness=-1

\subsection{Interaction Suitability}
PhysID initiates by assessing user-provided images using a MLLM. This  acts as a pre-filter, determining whether an image contains an object suitable for touch-based interaction on a mobile display. We leverage a pre-trained MLLM in a zero-shot setting, eliminating the need for task-specific data, annotations, or training. We explored various prompting strategies, including zero-shot, few-shot with text, Chain-of-Thought, and Few-shot Chain-of-Thought. The carefully designed prompt steers the MLLM towards generating an accurate classification label adhering to a specific format. Once an image is classified as interactable, a segmentation step extracts the primary object from the background \cite{kirillov2023segment}.\looseness=-1

\subsection{Implicit Characteristics Understanding}
Following image segmentation, PhysID employs the MLLM to categorize the object as either a soft or rigid body.  Soft bodies exhibit deformable behavior, akin to materials like cloth, rubber etc. Rigid bodies are objects that do not deform under normal conditions, maintaining their shape and volume.  For soft bodies, accurate simulation requires the inclusion of several material properties, with a few primarily governing motion simulation, such as Linear Stiffness, Damping Coefficient, Angular Stiffness, Volume Preservation, and Dynamic Friction Coefficient. Traditionally, these properties are determined through manual calibration for aesthetic and realistic simulation. However, this process is time-consuming, and these values can often be approximated within a specific range based on the object type, yielding similar rendering results. We exploit the MLLM's visual understanding through effective prompting to establish correspondences between objects in the image and their material properties. We eliminate manual calibration while achieving visually appealing and physically plausible soft-body simulations. \looseness=-1

In the case of rigid bodies, assuming the entire object to be rigid, its state can be described by a single position, orientation, and velocity. To enable subtle oscillatory motions on touch, push, and drag, we treat the complete object as a unified rigid body and apply hinge and cone twist constraints. These constraints allow an object to oscillate around a hinge while remaining restricted within a conical range of motion. While this technique is not universally applicable to all rigid bodies, our MLLM classifier effectively identifies suitable candidates. In PhysID, handling articulated rigid body dynamics (of connected parts) is beyond our current scope due to the complexity of segmenting and managing individual part movements within the generated mesh.\looseness=-1

\subsection{3D Mesh reconstruction}
Based on the MLLM's classification of the image as containing an interactive object, the Large Reconstruction Model (LRM) generates a 3D mesh representation of the object from the segmented image. For generating 3D representation, PhysID utilizes an off-the-shelf pre-trained model called Convolutional Reconstruction Model  (CRM) \cite{wang2024crm}. Given a single input image, CRM employs a multi-view diffusion model to generate six orthographic images and canonical coordinate maps(CCMs).  By leveraging Flexicubes \cite{shen2023flexible} as a geometry representation, CRM trains the reconstruction model in an end-to-end manner, using a combination of MSE loss and LPIPS loss for texture rendering during training.\looseness=-1

\subsection{Physics Engine and Interaction Simulation}
PhysID integrates an on-device physics engine for simulation. The  pipeline comprises three fundamental stages: collision detection, forward dynamics, and numerical time integration.\looseness=-1

Collision Detection: During this stage, the engine determines potential collisions between objects. A common approach utilizes a positive distance to indicate non-overlapping objects and a negative distance to signify intersections. For simplicity, PhysID treats each object as a single body, excluding multi-body collision scenarios.\looseness=-1

Forward Dynamics: This stage calculates the forces, inertia, and resulting accelerations acting on the rigid or soft bodies. Forces include friction, constraints imposed, velocity-dependent forces (e.g., drag), and position-dependent forces (e.g., springs). Newton's second law (F = ma) forms the basis for calculating the linear acceleration (a) of an object with mass (m) subjected to a force (F).  The engine computes the effect of an impulse on a mesh using,
\begin{equation}
\begin{aligned}
  \text{Impulse} &= F \Delta t = m \Delta v \\
  \Delta v &= \frac{\text{Impulse}}{m} &\text{;}& \quad \Delta \omega  = \frac{r \times \text{Impulse}}{I}
\end{aligned}
\label{eq:impulse}
\end{equation}
where $\Delta v$ is a linear component and $\Delta \omega$ is an angular component that involves computing the cross-product between the impulse vector and the relative position vector of the contact point to the center of mass, $r$, and $I$ depicts inertia of the object.\looseness=-1

Numerical Time Integration: The simulations necessitate the approximation of continuous motion, encompassing both velocity and position, across discrete time steps. Starting from initial conditions $v_{t_0}=v_0$ and $x_{t_0}=x_0$, the integration process accumulates the infinitesimal changes in these quantities over time. An efficient method to perform this integration is the use of implicit schemes. One widely adopted algorithm in real-time physics engines, is the symplectic Euler method \cite{muller2008real} \cite{coumans2015bullet}, defined as follows:
\begin{equation}
\begin{aligned}
    v_{(t+\Delta t)} &= v_t + a\Delta t  = v_t + \frac{F_{\text{ext}}}{m} \Delta t + \frac{\text{Impulse}}{m} \\
    x_{(t+\Delta t)} &= x_t + v_{(t+\Delta t)}\Delta t 
\end{aligned}
 \label{eq:velocity}
\end{equation}

\begin{table}[t]
\centering
\resizebox{\linewidth}{!}{
\begin{tabular}{l|lll|lll|lll|lll}
\hline
\multirow{2}{*}{\textbf{MLLM}} &

  \multicolumn{3}{c|}{\textbf{Zero Shot}}&
  \multicolumn{3}{c|}{\textbf{Few Shot}}&
  \multicolumn{3}{c|}{\textbf{Chain of Thought}}&
  \multicolumn{3}{c}{\textbf{Few Shot w/ CoT}}\\ \cline{2-13} 
                          &        T1& T2&T3& T1& T2&T3& T1& T2&T3& T1& T2&T3\\ 
                          \hline
{Honeybee \cite{cha2024honeybee}} 
                               & 0.51& 0.75&0.45& 0.75& 0.67&\textbf{0.69}& 0.62& 0.69&0.51& \textbf{0.75}& \textbf{0.76}&0.67\\ \hline
{LLaVa \cite{liu2023improvedllava}} 
                             & 0.31& 0.76&0.67& 0.63& 0.76&0.31& 0.64& 0.75&0.57& 0.56& 0.76&0.65\\  \hline
{CogVLM \cite{wang2023cogvlm}}   
                            & 0.74& 0.70&0.67& 0.72& 0.54&0.37& 0.56& 0.70&0.51& 0.60& 0.47     &0.67\\  \hline
\end{tabular}
}
\caption{Evaluation of MLLMs for classification tasks}
\label{tab:openmllm_classification}
\end{table}

External forces $F_{ext}$ can be incorporated into the system if they cannot be converted to positional constraints. In the simulation algorithm, velocities are dampened before being used to predict new positions. Finally, the velocities of colliding vertices are adjusted according to friction and restitution coefficients. Based on these intrinsic properties and a time step $\Delta t$, the dynamic behavior of the object is simulated. Due to its comprehensive capabilities we employ Bullet Physics \cite{coumans2015bullet} as the constraint solver module. \looseness=-1

\subsection{Sematic-aware soft-body dynamics simulation}
While conventional approaches rely on predefined material properties, we propose a novel method that leverages image segmentation, MLLM, and physics engine to achieve semantically varying behavior in soft bodies. Our approach allows certain regions of the mesh to remain static, while others exhibit dynamic behavior. We capture a snapshot of the full soft-body mesh rendered on a mobile device. We apply semantic segmentation \cite{kirillov2023segment} to this image; for each segmented mask, we query MLLM to determine whether it should exhibit static or non-static behavior. By leveraging this information, we create a binary mask that distinguishes static and non-static pixels in the screen/display resolution image. By mapping pixels from the mobile display to nodes in the mesh, we identify which nodes correspond to static pixels. Nodes associated with static pixels are set to have zero mass, effectively making them immovable. This approach offers several advantages, like avoiding complex mesh modifications required for static and dynamic regions. Further, eliminates the need for 3D segmentation by effectively utilizing 2D semantic segmentation and pixel to mesh-node resolution capability of the Physics engine.\looseness=-1

\begin{table}[t]
\centering
\resizebox{\columnwidth}{!}{

\begin{tabular}{cl|cccc}
\hline
\multicolumn{1}{l}{\cellcolor[HTML]{FFFFFF}\textbf{MLLM}} &
  \cellcolor[HTML]{FFFFFF}\textbf{Metric} &
  \cellcolor[HTML]{FFFFFF}\textbf{Zero Shot} &
  \cellcolor[HTML]{FFFFFF}\textbf{Few Shot} &
  \cellcolor[HTML]{FFFFFF}\textbf{CoT} &
  \cellcolor[HTML]{FFFFFF}\textbf{Few Shot CoT} \\ \hline
                                    & $w\text{-MSE}$ & 1.08E-02 & 6.73E-03 & 2.70E-03 & 4.87E-03          \\
\multirow{-2}{*}{\textbf{CogVLM} \cite{wang2023cogvlm}}   & $w\text{-MAE}$ & 3.18E-02 & 2.64E-02 & 1.59E-02 & 2.13E-02          \\ \hline
                                    & $w\text{-MSE}$ & 3.30E-03 & 1.99E-03 & 1.56E-03 & 1.96E-03          \\
\multirow{-2}{*}{\textbf{HoneyBee} \cite{cha2024honeybee}} & $w\text{-MAE}$ & 1.62E-02 & 2.13E-02 & 1.11E-02 & 1.26E-02          \\ \hline
                                    & $w\text{-MSE}$ & 5.07E-03 & 4.18E-03 & 2.88E-03 & 2.27E-03          \\
\multirow{-2}{*}{\textbf{LLaVa} \cite{liu2023improvedllava}}    & $w\text{-MAE}$ & 1.98E-02 & 1.32E-02 & 1.89E-02 & 1.17E-02          \\ \hline
                                    & $w\text{-MSE}$ & 5.86E-03 & 3.50E-03 & 2.17E-03 & \textbf{1.31E-03}          \\
\multirow{-2}{*}{\textbf{Gemini} \cite{team2023gemini}}   & $w\text{-MAE}$ & 2.28E-02 & 2.07E-02 & 1.26E-02 & \textbf{8.71E-03} \\ \hline
\end{tabular}
}
\caption{Evaluation of MLLMs for the softbody properties prediction against different prompting strategies}
\label {tab:physicsPropertyErrorEstimation}
\end{table}

\section{Experiments}
To rigorously evaluate  the PhysID framework, we conduct a comprehensive assessment using a diverse set of single-view images. This evaluation entails the selection of suitable images, automated assessment of their interaction potential within PhysID, generation of 3D object representations, and decoding of physical properties for accurate motion simulation and rendering. To evaluate the performance of each stage, we meticulously curated a testset and prepared annotations.\looseness=-1

\textbf{Dataset.} A curated subset of the Objaverse dataset \cite{deitke2024objaverse} serves as the foundation for evaluating the PhysID framework. From an extensive collection of 10 million objects, we sampled 5,000 objects belonging to the ‘nature-plants’ category, and 3,000 from the ‘fashion-style’ category. Subsequently, we further refined our selection, shortlisting a total of 285 objects that are suitable for PhysID. This subset includes 190 interactable and 95 non-interactable objects, with 98 objects under soft body dynamics and 92 with rigid body dynamics.\looseness=-1

\textbf{Data Annotation.}  For MLLM-based classification tasks, we established specific annotation guidelines to ensure consistency and accuracy in our dataset.  Interactable objects under rigid body dynamics are defined as those that can be hinged at the bottom and exhibit oscillatory motion when dragged, such as plants, flowers, etc. For soft body dynamics, we identified objects that behave with cloth-like motion, including shirts, fabric, flags, large leaves, and clothes.  Negative samples comprised diverse images including pets, large trees, heavy objects lacking oscillation, and humans wearing clothes where the expected motion arises from human activity rather than inherent soft body dynamics. 

For material property annotation, a data-driven approach with manual calibration was used to ensure realistic soft-body simulations on-device. All five physical properties were adjusted based on iterative simulations directly on the mobile device. Touch interactions and observations of object behaviors helped determine acceptable value ranges for each property. Additionally, the images of 3D objects were captured from angles maximizing viewability for generating 3D meshes.

\subsection{Experiments and Results}
\label{sec:MMLMExperiments}
We assess three open-source MLLMs, including CogVLM \cite{wang2023cogvlm}, HoneyBee \cite{cha2024honeybee}, LLaVa \cite{liu2023improvedllava} using our testset, focusing on,
\begin{enumerate}
\setlength\itemsep{-0.4em}
\item {Interactivity  Classification (T1):} Classify the
suitability of an object for interaction.
\item {Object Dynamics Classification (T2):} 
Classify the dynamics of an object as either rigid/soft body.\looseness=-1
\item {Mesh Semantics classification (T3):}
Classify the segmentation masks of soft body object as static/non-static.\looseness=-1
\item {Physics Property Estimation (T4):} Estimate the object's physical properties (e.g., stiffness, damping etc.).
\end{enumerate}

\begin{table}[t]
\centering
\resizebox{0.8\columnwidth}{!}{
\begin{tabular}{cllccc}
\hline
\textbf{Method}       &   \textbf{L1} $\downarrow$ & \textbf{L2} $\downarrow$&\textbf{PSNR} $\uparrow$ & \textbf{SSIM} $\uparrow$  & \textbf{LPIPS} $\downarrow$ \\
\hline
{TripoSR \cite{tochilkin2024triposr}}       &   0.055
&0.025
&16.10
& 0.79
& 0.207\\
{One-2-3-45 \cite{liu2024one}}    &   0.062
&0.032
&13.25
& 0.80
& 0.232\\
{LGM \cite{tang2024lgm}}&   0.56
&0.029
&12.98
& 0.78
& 0.211\\
{Instant Mesh \cite{xu2024instantmesh}}&   0.053
&0.030
&16.54
& 0.82
& 0.203
\\
\rowcolor{gray!25}\textbf{CRM \cite{wang2024crm}}           &   0.050
&0.026
&16.58
& 0.81
& 0.205
\\
\hline
\end{tabular}
}

\caption{Comparison of SOTA 3D mesh generation methods}
\label{tab:meshGeneration}
\end{table}

The implementation of prompt chaining with cloud-based inference of MLLMs forms the backbone of our end-to-end pipeline. An image classified as suitable for interaction automatically triggers the classification of its dynamics (soft/rigid). Subsequently, if the object exhibits soft body dynamics, we initiate the regression task using MLLMs to predict the physics properties and static-nonstatic regions of the image. Concurrently, a 3D model is constructed from the image using a cloud-hosted CRM \cite{wang2024crm}. The complete pipeline inference takes $\le $50s per image.\looseness=-1

To circumvent the need for labeled datasets and task-specific training, we leveraged the inherent knowledge within the MLLMs. We employ a zero-shot prompting approach, initially outlining the task with minimal detail and progressively increasing the information provided. Different prompting methods were explored, including few-shot prompting, chain-of-thought (CoT) prompting, and a combination of both. Given that the MLLMs employed for these tasks lack the capacity to accept multiple image tokens as input prompts, we leverage image captions generated by the same MLLM alongside the query image. Specifically, we incorporated ten diverse examples in the few-shot prompt setting for each classification task.  Following this, CoT prompting provided rationales for each example to guide the MLLM's reasoning process. We quantify the performance of T1 and T2 tasks using the F1-score.\looseness=-1 

Table \ref{tab:openmllm_classification} assesses three classification tasks (T1, T2, T3).  It can be observed that Honeybee \cite{cha2024honeybee}, fine-tuned on the ScienceQA dataset \cite{lu2022learn}, outperformed other models, suggesting that science-specific fine-tuning can directly influence visual grounding capabilities for understanding object dynamics. The few-shot CoT approach, providing a breakdown of classification examples with rationales, yielded a better F1 score compared to zero-shot. The zero-shot strategy exhibited a larger gap between precision and recall, due to an inability to correctly classify negative samples. Based on these observations we selected Honeybee\cite{cha2024honeybee} as our MLLM for classification and adopted the few-shot with CoT prompting strategy.\looseness=-1 

We compute weighted Mean Squared Error (w-MSE) and weighted Mean Absolute Error (w-MAE) for task T4 acknowledging the varying importance of different soft body properties. In Table \ref{tab:physicsPropertyErrorEstimation}, we observe two primary trends. First, incorporating additional context, examples, and rationales within the prompts leads to a notable improvement in MLLM performance, as evidenced by the decrease in error rates. Second, the few-shot CoT prompting strategy consistently yields strong performance across all MLLMs, with Gemini demonstrating the lowest error rates and CogVLM exhibiting the highest errors, indicating its limited suitability for this specific task.

In Table \ref{tab:meshGeneration} state-of-the-art 3D mesh generation models are evaluated on our dataset of interactive elements to assess reconstruction quality. CRM \cite{wang2024crm} achieved the highest PSNR (indicating high fidelity), while InstantMesh \cite{xu2024instantmesh} exhibited the best performance in terms of structural similarity. We chose CRM as our reconstruction model, considering the mesh quality of the generated model and on-device rendering.\looseness=-1

\begin{table}[t]
\centering
\resizebox{\columnwidth}{!}{
\begin{tabular}{ccccc}
\hline
\multirow{2}{*}{\textbf{Method}} & \multirow{2}{*}{\textbf{Real-time}} & \textbf{Dynamic}  & \textbf{Single-view} & \multirow{2}{*}{\textbf{Generalization}} \\
                        &                            & \textbf{Response} & \textbf{Image}  &                \\ \hline
DragAnything \cite{wu2024draganything}         &  $\times$  & $\times$ & $\times$ &  $\checkmark$             \\
ModalNeRF \cite{petitjean2023modalnerf}        &  \checkmark & $\times$  & $\times$ & $\times$               \\
PhysGauss \cite{xie2024physgaussian}        & \checkmark & \checkmark & $\times$ &  $\times$              \\
PhysDreamer \cite{zhang2024physdreamer}             & \checkmark & \checkmark&  $\times$ & $\times$           \\
GenerativeDynamics \cite{li2024generative}     &  \checkmark & $\times$ & \checkmark & $\checkmark$        \\
\rowcolor{gray!25}\textbf{PhysID (Ours)}            & \checkmark & \checkmark &  \checkmark & \checkmark      \\ \hline
\end{tabular}
}

\caption{Comparison of capabilities of different image dynamics modeling methods}
\label{tab:featureComparison}
\end{table}

Table~\ref{tab:featureComparison} provides a comparative analysis of various approaches to image dynamics. DragAnything \cite{wu2024draganything} employs trajectories for motion control, facilitating video generation. However, this restricts its suitability for real-time interaction. Conversely, ModalNerf \cite{petitjean2023modalnerf} enables real-time interactions but requires video as input, limiting its flexibility. While both PhysGauss \cite{xie2024physgaussian} and PhysDreamer \cite{zhang2024physdreamer} offer dynamic, real-time simulations, they require per-scene model training, hindering scalability. In contrast, PhysID achieves real-time, dynamic, and interactive simulations. It generalizes to a broad spectrum of objects using only single-view user-provided images, demonstrating superior scalability, however, texture quality relies on LRM's capability. Please check the project page for demo videos and detailed prompts.

\section{Conclusion}
PhysID introduces a training-free framework that enables physics-based interaction with static images on mobile devices. By leveraging MLLMs, it accurately classifies rigid and soft bodies, applying appropriate simulation techniques. For soft bodies, the framework estimates material properties and performs selective dynamic simulation without manual calibration. With CRM facilitating efficient mesh generation from single-view images, PhysID seamlessly integrates material properties and meshes into a physics-based engine for interactive simulation. PhysID establishes itself as a pioneering framework, streamlining the process of transforming static images into interactive environments on mobile devices.

\bibliographystyle{IEEEbib}
\bibliography{refs}

\end{document}